# Hybrid Self-Attention NEAT:
# A novel evolutionary approach to improve the NEAT algorithm

Saman Khamesian
s.khamesian@gmail.com

Hamed Malek
h_malek@sbu.ac.ir

## ABSTRACT

This article presents a "Hybrid Self-Attention NEAT" method to improve the original NeuroEvolution of Augmenting Topologies (NEAT) algorithm in high-dimensional inputs. Although the NEAT algorithm has shown a significant result in different challenging tasks, as input representations are high dimensional, it cannot create a well-tuned network. Our study addresses this limitation by using self-attention as an indirect encoding method to select the most important parts of the input. In addition, we improve its overall performance with the help of a hybrid method to evolve the final network weights. The main conclusion is that Hybrid Self-Attention NEAT can eliminate the restriction of the original NEAT. The results indicate that in comparison with evolutionary algorithms, our model can get comparable scores in Atari games with raw pixels input with a much lower number of parameters.

## I.   INTRODUCTION

Data plays an important role nowadays and is essential to produce and evaluate software and models. On the other hand, data dimensions have gradually increased with the spread of information and have posed challenging problems for scientists. In machine learning, which researchers have welcomed today, some previous studies are not compatible with these types of changes. Deep neural networks, which can learn high-dimensional representations, have performed far better than other algorithms in many areas like computer vision [1, 2, 3, 4], speech processing [5, 6], and reinforcement learning [7, 8]. These models are too large and complex with hundreds of millions of parameters which require plenty of computational resources. On the other hand, their performance strongly depends on the network architecture and parameter configuration [9].

In recent years, many studies in deep learning have focused on discovering specialized network architectures that apply to specific problems. Although there is a great deal of variation between deep neural network architectures, no specific rule has been established for choosing between them [10]. Consequently, finding the right design and hyper-parameters is essentially reduced to a black box optimization process. These configuration settings are usually determined from previous studies because manual testing and evaluation is a wearisome and time-consuming process that requires experience and expertise [9, 10]. An alternative approach is using neuroevolution algorithms to aid deep neural networks in building the architectures and learning hyper-parameters [11].

One of the most popular algorithms in the field of neuroevolution is the NEAT algorithm, which was introduced in 2002 by Stanley and Miikkulaineny [12]. The method solves the problems encountered in previous algorithms called Topology and Weight Evolving Artificial Neural Network (TWEANN) [13] by elevating the evolutionary operations between individuals of different lengths, evolving networks by adding neurons or connections in each iteration and protecting structural innovations by organizing them in species [13]. Due to using direct encoding in the NEAT algorithm, it seems that problems with high-dimensional input space (such as image processing tasks) make it prone to generate vast and complex networks. In this way, proper architecture is either done for a long time or may not be optimal.

In 2009, the Hypercube-Based Encoding for Evolving Large-Scale Neural Networks (HyperNEAT) algorithm [14] was introduced to solve this problem with the help of a population of small networks named Connective Compositional Pattern-producing Networks (CPPNs) as an indirect encoding technique [15]. Furthermore, it is necessary to specify the network architecture before execution, which loses its positive peculiarities. Indeed, despite solving the NEAT problem to some extent, the evolution of the connections and hidden neurons are virtually eliminated. Although in the ES-HyperNEAT (an extension of HyperNEAT) [16], an attempt was made to change the neurons' coordinations in some way during the evolution, it still does not have enough flexibility to change its architecture. Research has also shown that the NEAT algorithm works better in relatively complex problems than the HyperNEAT [17].

In this paper, we use the self-attention technique to omit the limitation of the NEAT algorithm in high-dimensional tasks so that it is not necessary to use all the input data to build the network, but only with the important parts which can help to solve the problem. In addition to evolving the value of the weights in the NEAT by employing regular operations, we use the Covariance Matrix Adaptation Evolution Strategy (CMA-ES) algorithm [18, 19] to optimize the final network weights. As we will show, our new method, Hybrid Self-Attention NEAT, is an ideal extension of the original NEAT because not only can we make it usable in high-dimensional spaces problem, we can also enhance its performance with the help of hybrid method in evolving its weights. Our study aims to showcase the power of the NEAT algorithm in problems like Atari Games with the raw pixel as input with much fewer parameters in comparison with other algorithms. We make the codes available for reproducing our experiments, and we hope our work will encourage further research into the neuroevolution of self-attention models and reinvigorate interest in hybrid methods[1].

## II. BACKGROUND

### A. *Neuroevolution*

Neuroevolution is a branch of artificial intelligence that uses evolutionary algorithms to generate neural networks or find their parameters [11]. It is mostly applied to Atari games, controlling roots and artificial life [20]. It is known that the neural network is actually inspired by the biological structure of the human brain [21]. The human brain becomes more complex and somewhat bigger during its lifetime (compared to childhood), which does not exist in neural networks. As a matter of fact, the number of neurons and layers is constant, and which neurons are connected is predetermined.

In the 1990s, the first steps in the evolution of neural network structures called TWEANNs were taken. In this series of studies, topology of neural networks changed slightly; for example, a neuron or connection to the parent was added when a child was created. Ronald Edmund first used a genetic algorithm in 1994 to find the optimal weights of neural networks [22]. At the same time, Frédéric Gruau, in his doctoral thesis, first proposed using genetic programming to change the structure and parameters of neural networks and increase their complexity [23]. The first question that arose about the design of TWEANNs algorithms was how to encode networks using an efficient representation to use them in evolutionary algorithms.

In the design of TWEANNs algorithms, binary coding was the simplest type of encoding. In 1992, Dasgupta and McGregor used a traditional bit string representation to represent the connection matrix called Structured Genetic Algorithm (sGA). [24]. This method had two fundamental disadvantages. First, the adjacency matrix size is the square of the number of nodes. Thus, the representation grows up for a vast number of nodes. Second, using a linear string of bits to represent a graph structure makes it challenging to ensure that crossover yields helpful combinations. After the binary coding method did not receive much attention, the graph coding method became popular. In 1997, Pujol and Poli used a dual representation to modify mutation and crossover in their Parallel Distributed Genetic Programming (PDGP) algorithm. The first is a graph, and the second is a linear genome of node definitions to specify the incoming and outgoing connections [25].

---

[1] https://github.com/SamanKhamesian/Hybrid-Self-Attention-NEAT

*B. NEAT*

In 2002, Kenneth Stanley introduced a kind of neuroevolution approach called NEAT, and the purpose of designing it was to solve the problems of TWEANNs algorithms [12]. In addition to updating weights in the NEAT algorithm, the network architecture changes over generations and becomes more complex. Numerous studies have shown that it works well in board games, challenging control tasks, and decision-making problems [20, 26, 27]. The unique part of this algorithm is that it provides a new representation based on direct encoding to create networks and define new evolutionary operations. The method starts with a population of small networks without any hidden units. Through the evaluation, each network can grow by adding neurons or connections and grouped in different species to protect the structural innovations. In this way, networks with different architectures can be produced, and the most appropriate ones can be selected based on the desired evaluation criteria; Countless architectures and parameters need to be set before execution.

*C. HyperNEAT and CPPNs*

This section reviews the HyperNEAT algorithm, one of the most important extensions of the NEAT, developed by Stanley et al. in 2009 [14]. The algorithm takes the indirect coding approach to reduce the volume of NEAT in complex problems. HyperNEAT uses small networks named CPPNs [15] for determining the weights of its network called substrate and uses the NEAT algorithm to evolve them. As shown in Figure 1, the ANN substrate has an input layer consisting of neurons take on Cartesian coordinate $(X_1, Y_1)$. CPPNs that represent spatial patterns in hyperspace learn the problem space geometry and find the weights between the input layer and the next layer network containing nodes spanning an $(X_2, Y_2)$ plane. To compute all the weights in the network, we have to range the inputs of the CPPN over all pairs of nodes in the two adjacent (input and output) layers. Similar to ANNs, the substrate is determined before the program is executed, but network connectivity can take many forms, such as grid, circular or three-dimensional layers [14]. The HyperNEAT can proceed in five stages as follows:

1. First, you need to specify a configuration for the desired output pattern (substrate definition).
2. Produce a population of small CPPNs (without hidden neurons) with random weights.
3. Repeat steps 4 and 5 until you find the final solution:
4. For each member of the CPPN population:
   - All weights of the substrate have to be calculated by the chosen CPPN.
   - The generated substrate should be used as a final model to its fitness.
5. The NEAT method evolves CPPN population, and the new generations are created.

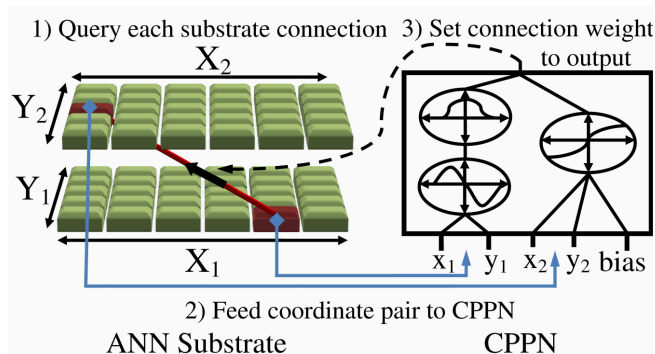

**Figure 1:** HyperNEAT evolution: The NEAT algorithm evolves the CPPNs (right). This CPPN has to set every connection weight in the substrate subsequently (left) (Figure is duplicated from [28])

*D. AE-NEAT*

In 2020, AE-NEAT [29] was introduced by Adam Tupper to present an extension of the NEAT algorithm and examine it in general video games [30]. Figure 3 shows the general architecture of the AE-NEAT algorithm, with the compressor section for the encoder and the controller section for the NEAT. At each time step, an observation, an image of the game screen, comes from the environment as an input. This image is converted to a compact state representation by the compressor. This compact state representation is then fed to the controller, designed to make a decision and produce the proper action for the game agent. This cycle continues until the end of the episode is reached. For the compressor section, two different auto-encoders are examined in this study, and the final one is shown in Figure 2. The trained encoder section extracts features and generates proper input for the controller.

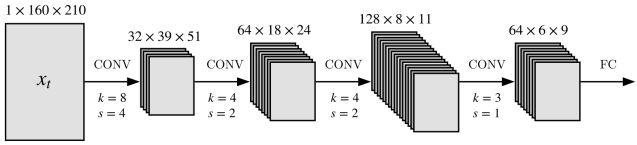

**Figure 2:** The architecture of the encoder which is used in the AE-NEAT algorithm. (Figure is duplicated from [29])

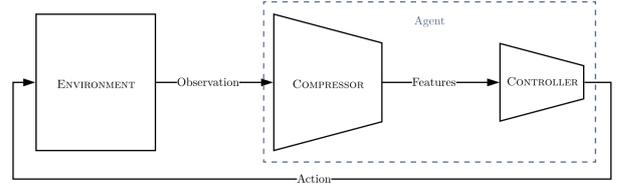

**Figure 3:** An abstract data-flow diagram of the AE-NEAT algorithm (Figure is duplicated from [29])

*E. Self-Attention*

Self-Attention is an attention mechanism relating various sorts of a sequence to calculate its representation and is the fundamental operation of any transformer. For the first time, transformers architecture was proposed in the paper "Attention is All You Need" by Google in 2017 [31]. In summary, transformers' idea is to compute its input and output representation using a self-attention layer instead of sequence-aligned RNNs or convolution. After that, transformers have mostly become more superficial, so that it is now much more straightforward to explain how modern architectures work. Self-Attention has been used in many areas like abstractive summarization [32], image generation [33], and sentence embedding [34].

## III. METHODOLOGY

*A. Input Preparation*

As mentioned, self-attention is applied to sequential data, but it is straightforward to modify it so that it could also be applied to images. In this work, we were inspired by one of the practical usages of self-attention in extracting information from raw images, which was developed by Yujin Tang and al. [35]. Instead of working on every pixel, they organize the input image into patches and use a simple variation of self-attention architecture to detect the important patches via an attention score matrix. We refer to [35] for an in-depth overview of the self-attention module used in this study.

*B. Self-Attention Parameter Training*

The self-attention module has Key and Query parameters that musct be calculated. Since we do not know the best values of the matrices (that is, we cannot determine what we are looking for) and we cannot determine the loss function, the best way to learn the values of these matrices is to use evolutionary methods. Among all the evolutionary algorithms, the CMA-ES algorithm has performed well in related studies [18, 19, 36], which is why we chose it.

First, we create a population of chromosomes that contain the parameters of the Key and Query matrices. Each of these chromosomes has a fitness that must be calculated. One solution is to use a pre-trained model to play the game as well as possible and get an acceptable score. In this case, we can run the CMA-ES algorithm, and in each iteration, selected chromosomes can determine the unknown values for the Key and Query matrices. Consequently, the self-Attention module can extract the crucial patches and give them to the pre-trained model to play the game. Finally, the model calculates the total reward, which is the fitness score for the selected chromosome.

However, we did not have a pre-trained model to play the game, and playing the game at random is also not a wise choice due to the lack of scenarios and images of different situations. This is why we use an idea to evolve both NEAT and self-attention parameters simultaneously and named it the *Seesaw method*. In this way, we fix the network by selecting the best member of the NEAT population as a candidate for controller section to play the game. Then, we select the members of the CMA-ES population one by one, to set the self-attention parameters (compressor section). The self-attention module prepares the proper input for the NEAT candidate to calculate the fitness for each of the members. After calculating all fitness values, we choose the best member to set the self-attention parameters for that current iteration. Next, we select all members of the NEAT population one by one to play the game and calculate their fitness. The whole process is summarized in Figure 4.

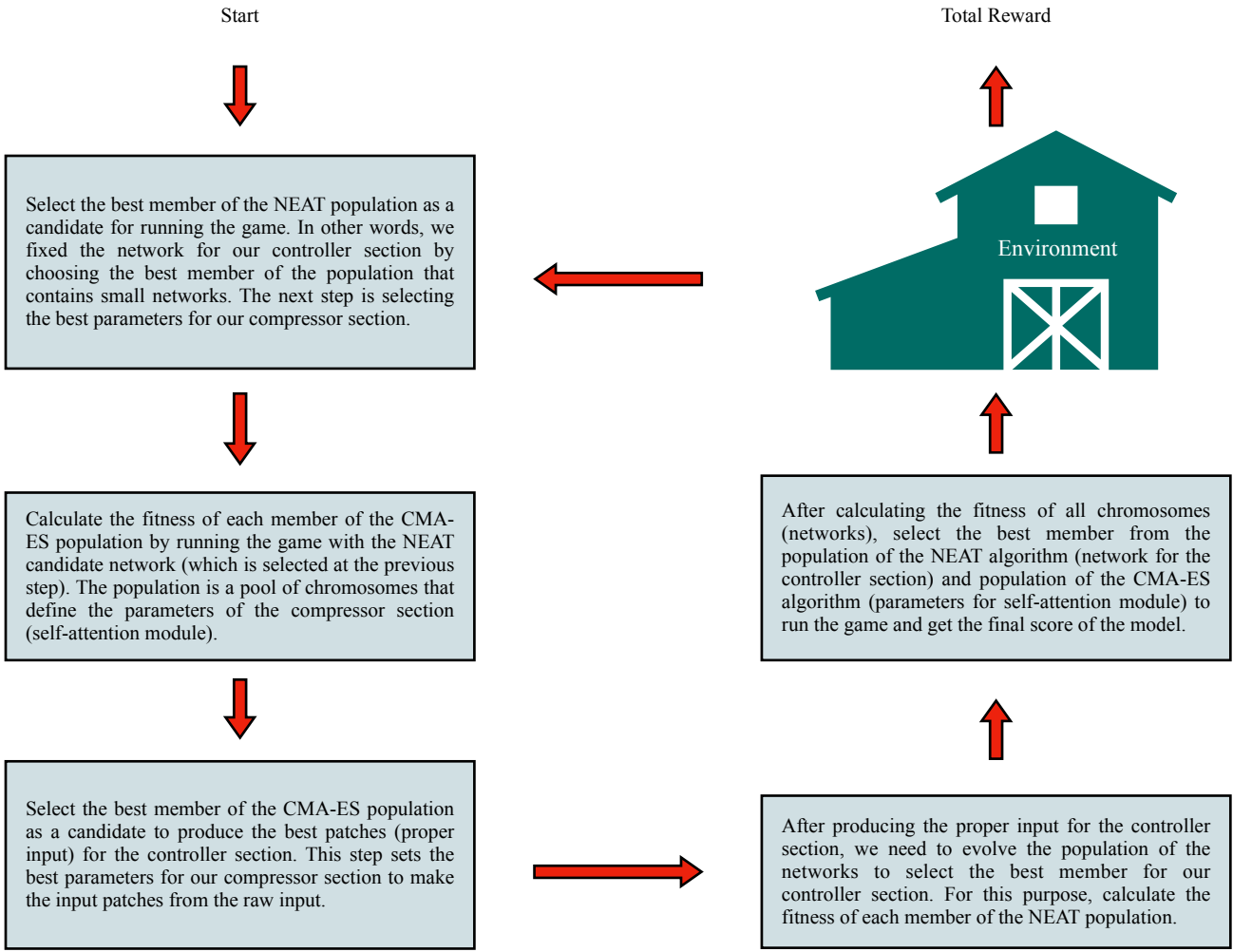

**Figure 4:** Flowchart of the whole process in one iteration

## C. NEAT Weights Tuning

Once the NEAT and self-attention learning cycle is complete, we can select the best member of the NEAT and CMA-ES populations (which have the highest fitness) as the final model. However, NEAT network weights may not have reached their optimal value. It should be noted that some constants are defined as the probability of mutation and crossover of weights in the NEAT algorithm. These parameters determine whether the values of the weights change or not. If yes, the weight should be defined numerically at random or swapped by another edge [12]. As can be seen, this method of updating weights may not lead to producing the optimal values. So we add another step to improve the weights differently. We fix the network topology and the self-attention parameters in this step to only the weights change. Since the topology of NEAT networks has backward connections in different ways, using gradient-based algorithms can sometimes be challenging, so the most reliable way to learn weights is to use evolutionary methods. Using several evolutionary algorithms (hybrid method) to optimize weights can improve network performance. While different evolutionary algorithms can be used for this purpose (like the genetic algorithm), we use the CMA-ES algorithm, which obtained very good performance in the previous works [18, 19, 36], and left the other ones to the enthusiasts of this study.

*D. Conclusion*

In the proposed method called Hybrid Self-Attention NEAT, we tried to segment the input data and identify the efficient parts for the agent by using the self-attention technique. As a result, the agent in the game becomes blind to the unnecessary parts of the input. We then used the CMA-ES algorithm to obtain the unknown parameters of the attention matrix. Simultaneous evolution of NEAT algorithm with optimization of self-attention parameters was another innovative way to fit the model. Finally, to optimize the weights of the final model, we used the hybrid method, and with the help of CMA-ES, we optimized the weights of the final NEAT network.

IV. EXPERIMENTS

One of the most popular reinforcements learning video game platforms is Arcade Learning Environment (ALE) [37]. ALE offers a standard reinforcement learning level for over 50 Atari 2600 games based on the Stella Emulator. Each game has a discrete movement space containing eighteen actions, including all possible input combinations in Atari 2600 games (each lever includes nine directions of movements and a fire button). Also, in some games, discrete space is limited to permissible movements within the game. ALE provides a variety of inputs for a game, including $210 \times 160$ pixels image input and 128 bytes of console RAM.

Another platform is the OpenAI Gym Environment which is used in this study. This platform is written with Python, and allows us to test various games with little complexity. The system used to test the agent has the following configurations: Intel® Xeon® CPU E5-4650 v2 2.4GHz, 16 core processor, and 24GB RAM. Due to the hardware limitations, we use the minimum possible values for the number of iterations, population size, and trials. A complete list of these parameters can be found in the Appendix. Also, we determine some hyper-parameters by default based on previous research. Obviously, by increasing the population size and testing different values of the hyper-parameters to choose the most appropriate ones, better results can be obtained. We entrust this part of the research to those interested in completing or continuing this study in the future.

It should be mentioned that we evolve a separate configuration for each game. This configuration consisted of 50 generations with a population size of 64 individuals for the NEAT and 32 individuals for the CMA-ES algorithm. Each member of the population earned a fitness score equal to the game score gained from a single episode of play (e.g., until the game-over or the frame limit is reached). Due to the randomness in this domain, we evaluate every individual of each generation three times and reported average scores to get a better estimate of the true elite. The diagram in Figure 5 shows the effect of this method. According to this chart, the fitness fluctuations are significantly reduced when the number of trials for each member is considered three compared to the case where the number of trials per member is one. In the following, we provide experimental results obtained from applying Hybrid Self-Attention NEAT on three Atari games, Asteroids, Berzerk, and Space Invaders.

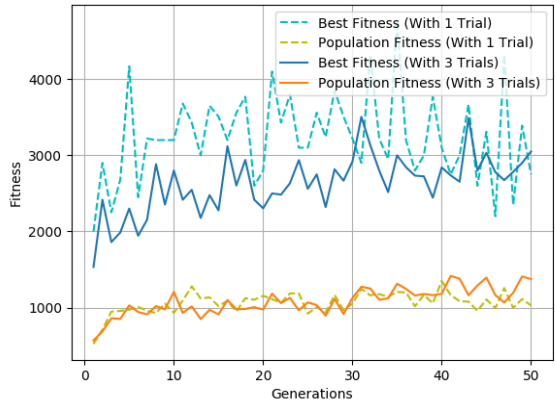

**Figure 5:** Comparison of two different performances of Asteroids - The number of times the game was played per iteration for each member of the population is one and three.

## A.  *Asteroids*[1]

In Asteroids, the player controls a spaceship that is periodically traversed by flying saucers. The object of the game is to shoot and destroy the asteroids and saucers without colliding or being hit by the saucers' counter-fire. The average fitness of the population and the best fitness throughout the first stage of the training process can be seen in Figure 6 (a). We tune the network weights via the CMA-ES algorithm in the second stage, shown in Figure 6 (b).

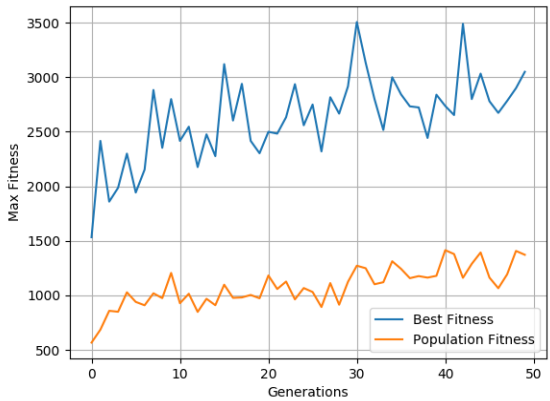
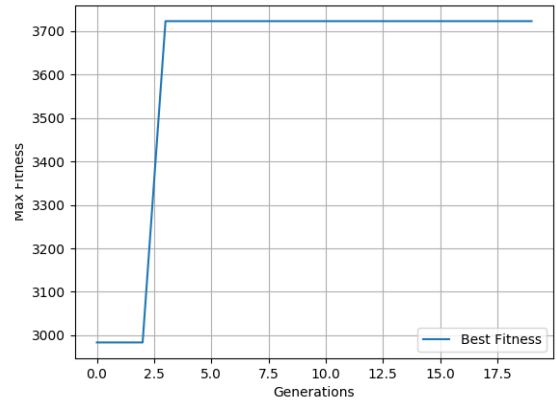

(a)                                                                 (b)

**Figure 6 (a):** First stage of the training session of the Hybrid Self-Attention NEAT algorithm for the Asteroids Atari game - The best and average fitness of the population is shown for 50 generation. **Figure 6 (b)** Second stage of the training session of the Hybrid Self-Attention NEAT algorithm for the Asteroids Atari game.

## B.  *Berzerk*[2]

In the Berzerk video game, the player controls a green stick man. Using a joystick for moving and a firing button for shooting, the player navigates a simple maze filled with many robots, who fire lasers back at the agent. A player can be killed by being shot, running into a robot, or coming into contact with the electrified walls of the maze itself. The average fitness of the population and the best fitness throughout the first stage of the training process can be seen in Figure 7 (a). The result of tuning network weights with the CMA-ES algorithm can be found in Figure 7 (b).

---

[1] https://www.gymlibrary.ml/environments/atari/asteroids/

[2] https://www.gymlibrary.ml/environments/atari/berzerk/

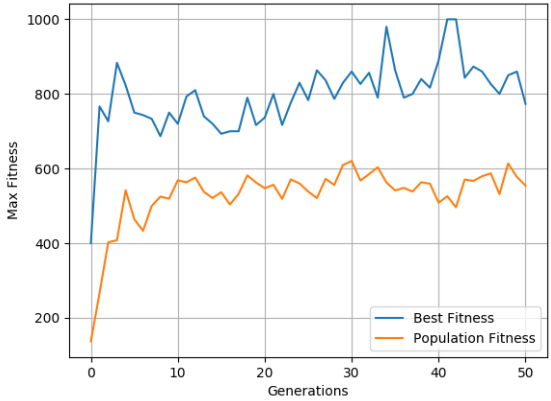
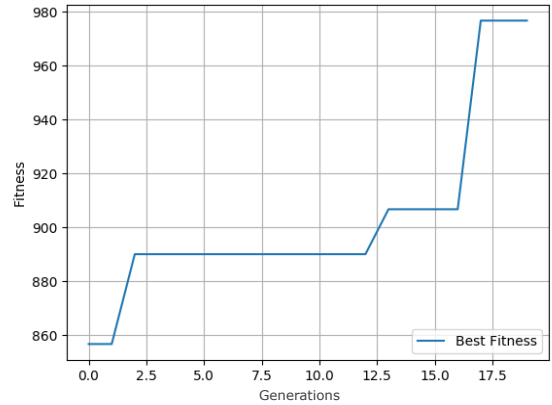

(a)                                                  (b)

**Figure 7 (a):** First stage of the training session of the Hybrid Self-Attention NEAT algorithm for the Berzerk Atari game - The best and average fitness of the population is shown for 50 generation. **Figure 7 (b):** Second stage of the training session of the Hybrid Self-Attention NEAT algorithm for the Berzerk Atari game.

### C.    Space Invaders[1]

Space Invaders is a two-dimensional fixed shooter game in which the player controls a laser cannon by moving it horizontally across the bottom of the screen and firing at descending aliens. The game's object is to defeat an alien and earn points by shooting it with the laser cannon. The result of the first and the second training sessions are shown in Figure 8.

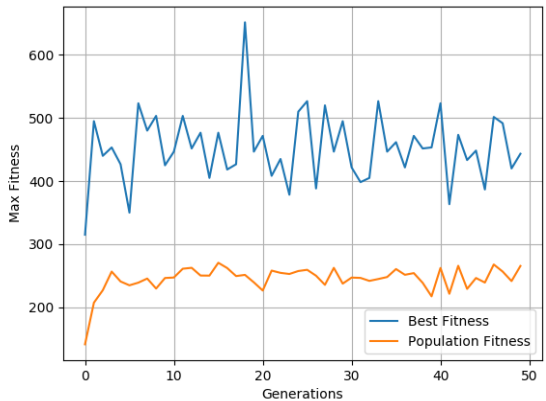
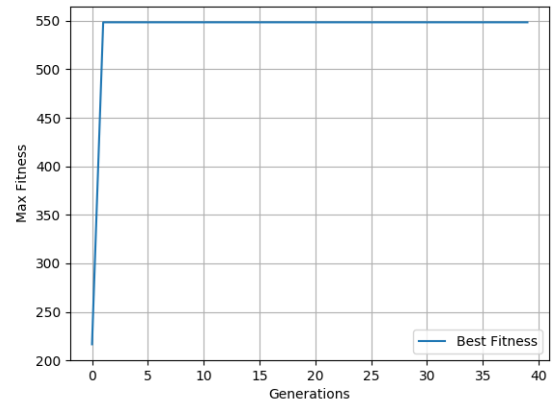

(a)                                                  (b)

**Figure 8 (a):** First stage of the training session of the Hybrid Self-Attention NEAT algorithm for the Space Invaders Atari game - The best and average fitness of the population is shown for 50 generation. **Figure 8 (b):** Second stage of the training session of the Hybrid Self-Attention NEAT algorithm for the Space Invaders Atari game.

### D.    Seaquest[2]

In Seaquest game, you control a sub able to move in all directions and fire torpedoes. The goal is to retrieve as many divers as you can, while dodging and blasting enemy subs and killer sharks; points will be awarded accordingly. The average fitness of the population and the best fitness throughout the first stage of the training process can be seen in Figure 9 (a). We tune the network weights via the CMA-ES algorithm in the second stage, shown in Figure 9 (b).

---

[1] https://www.gymlibrary.ml/environments/atari/space_invaders/

[2] https://www.gymlibrary.ml/environments/atari/seaquest/

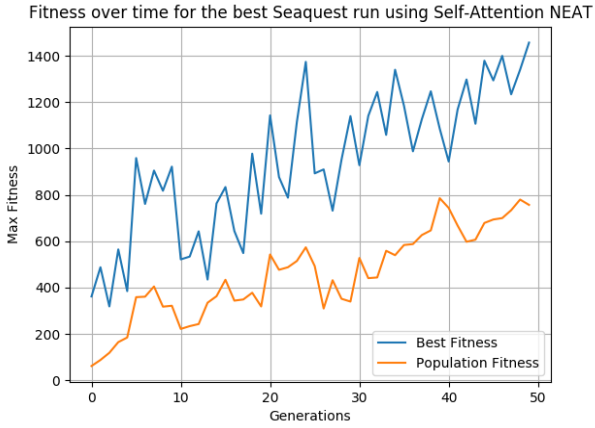
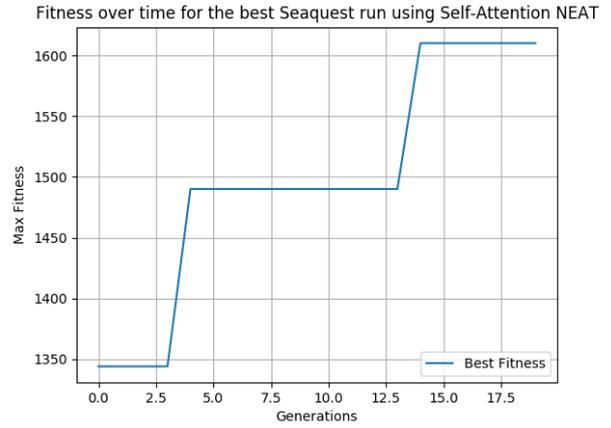

(a)                     (b)

**Figure 9 (a):** First stage of the training session of the Hybrid Self-Attention NEAT algorithm for the Seaquest Atari game - The best and average fitness of the population is shown for 50 generation. **Figure 9 (b):** Second stage of the training session of the Hybrid Self-Attention NEAT algorithm for the Seaquest Atari game.

The result of the self-attention module in preparing the input used to control the agent is also shown in Figure 10 for all three games. The ten most essential patches which were selected with the self-attention module are presented with white squares.

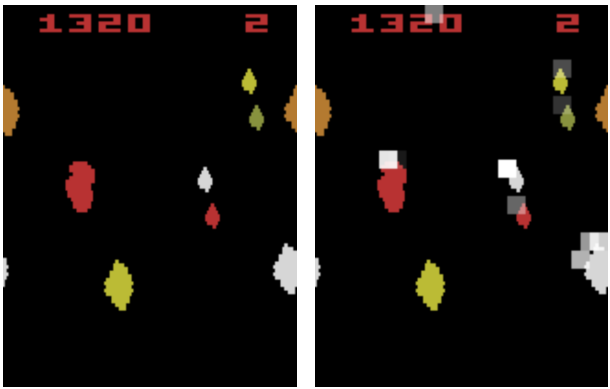
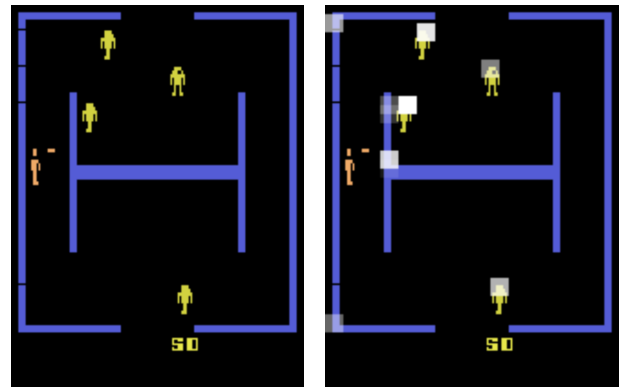

(a)                     (b)

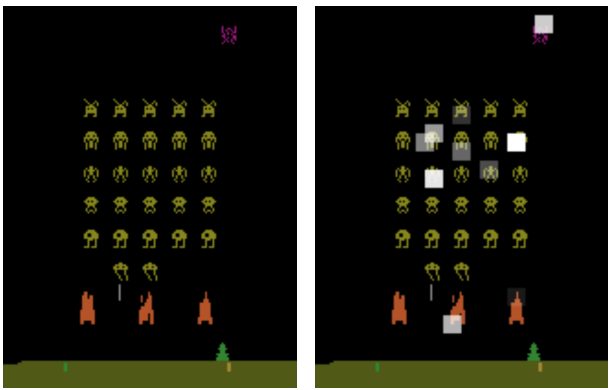
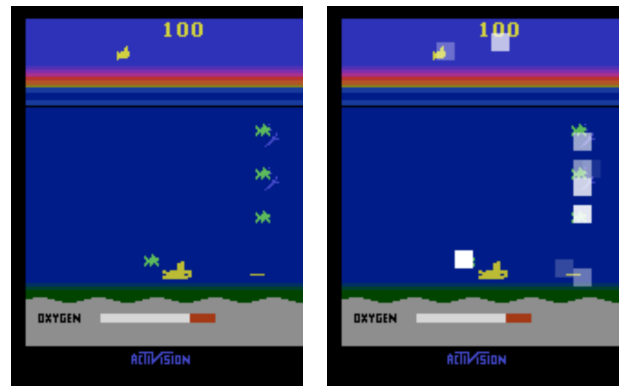

(c)                     (d)

**Figure 10:** A random screenshot of the environment of three Atari games (Asteroids, Berzerk, Space Invaders and Seaquest) (Left ones) besides the result of the Self-Attention module in detecting the most important patches (Right ones).

## E. Results

We report the average score of each game over 100 consecutive tests with standard deviation and compare them with other evolutionary algorithms on similar datasets, and pixel inputs include AE-NEAT, HyperNEAT, Deep GA, and OpenAI ES in Table 1 [26, 29, 38, 39]. We also compare our results against Agent57 [40], the highest performing general Atari game playing method of any type to date, and the expert human performance in all three Atari games supported by the ALE [8].

Table 1: Scores for our hybrid method compared against other evolutionary methods that learn from raw pixel inputs, a random agent, expert human performance, and Agent57 (the current state-of-the-art general Atari game playing agent).

| Game | Ours | HyperNEAT | AE-NEAT | OpenAI ES | Deep GA | Random | Human | Agent57 |
|---|---|---|---|---|---|---|---|---|
| Asteroids | **1832.37 (±752.4)** | 1694.0 | 1739.0 (±596.1) | 1562.0 | 1661.0 | 1013.7 (±439.0) | 13157.0 | 150854.6 |
| Berzerk | 926.67 (±247.34) | **1394.0** | 855.7 (±233.5) | 686.0 | - | 162.0 (±118.1) | 2237.5 | 61507.8 |
| Space Invaders | 625.0 (±258.88) | **1251.0** | 729.9 (±189.3) | 678.5 | - | 157.4 (±101.0) | 1652.0 | 48680.9 |
| Seaquest | **1523.61 (±220.6)** | 716.0 | 630.2 (±126.8) | 1390.0 | 798.0 | 88.9 (±64.3) | 20182.0 | 999997.6 |

The NEAT algorithm is not included in this comparison because it does not apply to pixel input in principle [26]. In Asteroids and Seaquest, our proposed model has the best performance compared to other algorithms, and in Berzerk, it has been better than AE-NEAT. However, due to hardware limitations, we do not consider optimal parameters, sufficient population, and the appropriate number of iterations for Hybrid Self-Attention NEAT.

With the help of the self-attention technique, we can significantly reduce the total number of parameters. As shown in Table 2, our model has 300x fewer parameters than the HyperNEAT model. The exact number of the connections used in the final policy network (for example, NEAT) differs from game to game, so we report an approximate number for the total number of parameters. Also, according to AE-NEAT architecture, the exact number of parameters used in the encoder is equal to 239,904, while the number of learnable parameters of the Hybrid Self-Attention NEAT model is about 100 times fewer (2408 parameters except the number of connections in the final network). It should also be noted that the encoder used in the AE-NEAT algorithm must be trained before execution. To learn that AutoEncoder, we need multiple images of the game's environment, and to get these images, a pre-trained model must run the game many times. Accessing such a model and training it with hundreds of thousands of images has its own challenges [29]. However, in the proposed Hybrid Self-Attention NEAT model, we train the self-attention module and the NEAT algorithm in a simultaneous repetition loop, and there is no need for separate training, extracting multiple images from different games manually, and a pre-trained model.

Table 2: Comparison of the total number of parameters used in each model with Hybrid Self-Attention NEAT (Ours):

| Model Name | Total Parameters |
|---|---|
| HyperNEAT | ~907,000 |
| AE-NEAT | ~240,000 |
| Deep GA | ~4,000,000 |
| OpenAI ES | ~120,000 |
| **Hybrid Self-Attention NEAT (Ours)** | **~3000** |

## V. DISCUSSION

As mentioned earlier, the purpose of this study is to improve the NEAT algorithm in the visual data set, which due to the large input size, the algorithm could not run properly. Also, the effect of the proposed combined method to optimize the final weights of the model was examined in the result section.

A few points to keep in mind about the Atari game dataset, one of the most popular datasets on reinforcement learning issues: First of all, it should be noted that different games have their own rules and challenges, and the information obtained from the game environment images is unique to each game. For example, in Go [41, 42], all game information can be extracted in one frame, but there are other games that it is too difficult to produce the right action for the agent by taking only one frame from the game environment. Of course, in addition to images, there is another way to get game information by APIs that only include the coordinates of objects and game elements (RAM information), but these inputs are out of our discussion.

Here we address the Car Racing[1] game to get a better understanding of this issue. The input of this game is $96 \times 96$ pixel frames with three RGB channels. In this game, the agent (racing car) must complete one round of the race track. The track map is not fully available, and only a part of it can be displayed as Bird View in each frame (Figure 11). Along the way, there are random checkpoints where a positive score is recorded when the car passes. The location of these checkpoints is different in each simulation, but they all cover the width of the road, and if the car moves off-track (for example, off-road), it cannot receive these points. Also, by default, -0.1 points are assigned to the agent per frame. This game ends in a maximum of one thousand frames, and if the score is more than 900, it wins. The model that wants to control this factor must produce three outputs: gas, brake, and direction of movement (straight, left, and right). The numerical gas and the brake are between zero and one, and the direction is between -1 to 1. The complexity of this game is that since it has only one frame, it is not possible to get the speed or direction of the car, and therefore the condition of the car is indescribable for the model. Therefore, only models that have memory and can extract such information from a sequence of several frames can win the game [35]. Although the NEAT network can support backward connections, it does not have a memory unit and cannot succeed in such problems, unlike Long Short-Term Memory (LSTM) networks [43]. All the models that set a record in this game [35, 44, 45] used the LSTM model. The result of running Car Raring using the Self-Attention NEAT algorithm presented in Figure 12 tells us that the minimum acceptable score in this game cannot be reached without a memory network.

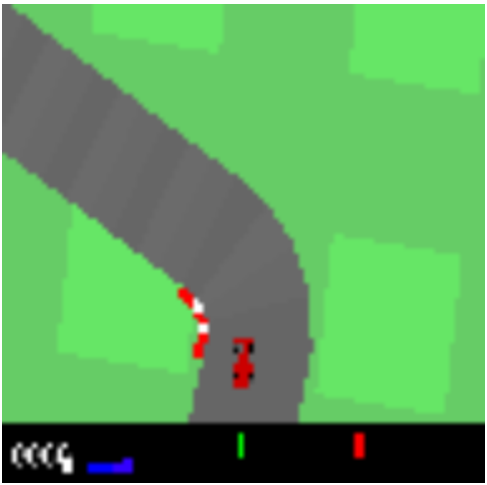

**Figure 11:** Sample frame of Car Racing game

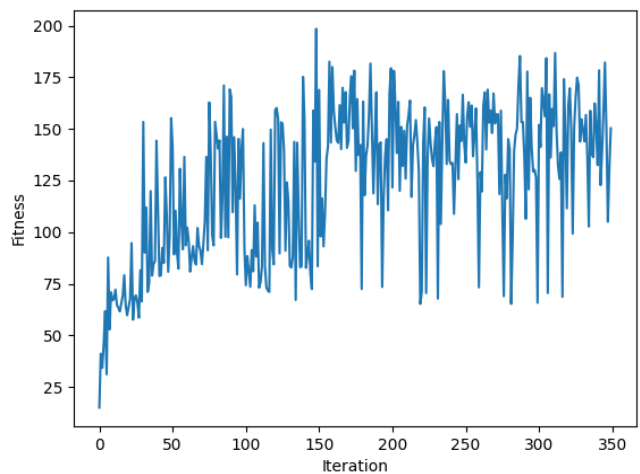

**Figure 12:** Fitness of the Self-Attention NEAT algorithm for the Car Racing Game

---

[1] https://www.gymlibrary.ml/environments/box2d/car_racing/

## VI. CONCLUSION

One of the improvements made to the NEAT algorithm was updating the weights, which we introduced as a hybrid method, and in the result section, we showed that this hybrid method using CMA-ES could significantly affect this method's network performance. On the other hand, using the self-attention technique, we significantly reduced the total number of learnable parameters of the model so that the NEAT algorithm can be efficiently run on high-dimensional problems. Our model can achieve comparable results to the same algorithms on Atari games with raw pixel input, and specific in Asteroids and Seaquest outperform evolutionary methods.

Resources limitations meant that the variety of experiments in this study was not as large as it should be and did not show the maximum potential of this model. Therefore we can address the below items for future studies:

- **Variety of games and use of different datasets:** Since the proposed model has the ability to generalize the problem, it is possible to test visual datasets with which the NEAT algorithm is compatible. Also, the variety of Atari games can be improved, and the results can be compared by categorizing them (for example, in terms of game style).

- **Testing of various parameters:** Some essential parameters must be determined before running, and, as mentioned earlier, these parameters are based on previous research and experiments. Since the NEAT network can be susceptible to these parameters, one way to improve the results is to use optimal parameters. Among the parameters that were directly affected during the various experiments of this study, the following can be mentioned:
  - Population Size
  - Activation Functions
  - Number of Iterations
  - Mutation and Crossover Probability
  - Number of Important Patches (K)

- **Experimenting with different feature extraction functions from patches:** As mentioned before, there are many functions to select the features from important patches. In this research, the simplest ones, namely mapping each patch to the coordinates of its center. But some other functions can be used to obtain the rich properties of these components and directly affect the model's performance. Functions that may themselves consist of learnable parameters.

# VII. APPENDIX

## A. Atari Games Policy Learning Experiments

Table A.1: Hyper-parameters for the NEAT algorithm

| Hyperparameter | Value |
|---|---|
| Population Size | 64 |
| Fitness Criteration | Max |
| Reset on Extinction | TRUE |
| Activation Function | Sigmoid |
| Aggregation Function | Sum |
| Compatibility Distance Disjoint/Excess Gene Coefficient | 1 |
| Compatibility Distance Weight Difference Coefficient | 0.4 |
| Add Connection Probablity | 0.05 |
| Delete Connection Probability | 0.05 |
| Feed-Forward | FALSE |
| Add Node Probability | 0.03 |
| Delete Node Probability | 0.03 |
| Weight Range | [-30.0, +30.0] |
| Weight Mutation Power | 0.05 |
| Weight Mutation Rate | 0.8 |
| Weight Replace Rate | 0.1 |
| Compatibility Distance Threshold | 3 |
| Max Stagnation | 15 |
| Species Elitism | 2 |
| Elitism Threshold | 5 |
| Survival Threshold | 0.2 |

Table A.2: Hyper-parameter for the CMA-ES algorithm

| Hyperparameters | Value |
|---|---|
| Population Size | 32 |
| Init Sigma | 0.1 |

Table A.3: Hyper-parameter for the Self-Attention part

| Hyperparameters | Value |
|---|---|
| Patch Size | 10 |
| Patch Stride | 5 |
| Transformation Dimension | 4 |
| Number of K (Top Selected Parches) | 10 |